\pdfoutput=1

\documentclass[11pt]{article}

\usepackage[]{EMNLP2023}

\usepackage{times}
\usepackage{latexsym}

\usepackage[T1]{fontenc}

\usepackage[utf8]{inputenc}

\usepackage{microtype}

\usepackage{inconsolata}
\usepackage{array}
\usepackage{times}
\usepackage{graphicx}
\usepackage{latexsym}
\usepackage{CJKutf8}
\usepackage{amsfonts}
\usepackage{amsmath}
\usepackage{mathtools}
\usepackage{multicol}
\usepackage{multirow}
\usepackage{color}
\usepackage{booktabs}

\title{Unified Lattice Graph Fusion for Chinese Named Entity Recognition}

\author{
    Dixiang Zhang$^\textnormal{1}$ \qquad
    Junyu Lu$^\textnormal{12}$ \qquad
    Pingjian Zhang$^\textnormal{1}$\footnotemark[2] \qquad
    \\
    $^\textnormal{1}$South China University of Technology \quad
    $^\textnormal{2}$International Digital Economy Academy \quad
    \\
    {\tt\small \{zhangdixiang, lujunyu\}@mail.scut.edu.cn \qquad zhangpingjian@scut.edu.cn}
}

\begin{document}
\maketitle
\begin{abstract}
Integrating lexicon into character-level sequence has been proven effective to leverage word boundary and semantic information in Chinese named entity recognition (NER). However, prior approaches usually utilize feature weighting and position coupling to integrate word information, but ignore the semantic and contextual correspondence between the fine-grained semantic units in the character-word space. To solve this issue, we propose a Unified Lattice Graph Fusion (ULGF) approach for Chinese NER. ULGF can explicitly capture various semantic and boundary relations across different semantic units with the adjacency matrix by converting the lattice structure into a unified graph. We stack multiple graph-based intra-source self-attention and inter-source cross-gating fusion layers that iteratively carry out semantic interactions to learn node representations. To alleviate the over-reliance on word information, we further propose to leverage lexicon entity classification as an auxiliary task. Experiments on four Chinese NER benchmark datasets demonstrate the superiority of our ULGF approach.
\end{abstract}

\section{Introduction}
Named Entity Recognition (NER) aims to identify entities from free text and classify them into pre-defined categories, such as PER, LOC and ORG~\cite{zhao2021modeling}. Compared with English NER~\cite{lample2016neural, ma2016end, sun2020learning}, Chinese NER is more challenging since there is no explicit word boundary. 

Traditional methods divide Chinese NER into two individual sub-tasks: word segmentation and word-level sequence tagging~\cite{yang2016combining, cao2018adversarial}, but the error propagation caused by incorrect segmentation will negatively impact the identification of named entities~\cite{peng2015named, he2017f}. Moreover, since word semantics and boundary information are potentially beneficial for character-level sequence learning, the lattice structure has been proposed to integrate matched lexicon information into each character~\cite{zhang2018chinese}. Prior approaches have attempted to design various network architectures compatible with lattice inputs in the character sequence, such as RNN-based~\cite{liu2019encoding}, Graph-based~\cite{gui2019lexicon}, CNN-based~\cite{gui2019cnn}, and Transformer-based~\cite{mengge2020porous,li2020flat} models, which integrate word information in a more effective and efficient way with relative position encoding and char-to-word attention weighting.


\begin{figure*}[t]
\begin{center}
\centerline{\includegraphics[width=0.95 \linewidth]{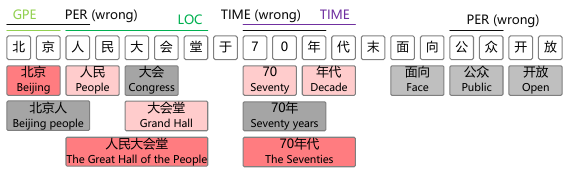}}
\vskip -0.1in
\caption{The character-words pair sequence of the truncated Chinese sequence \begin{CJK*}{UTF8}{gbsn}``北京人民大会堂于70年代末面向公众开放\ (The Great Hall of the People in Beijing was opened to the public in the late 1970s)''.\end{CJK*} Light/Dark red boxes represent effective words with low/high contribution. Gray boxes represent disturbed words.}\label{fig:example}
\end{center}
\vskip -0.4in
\end{figure*}

Despite the success, the existing methods have not considered the impact of contextual information on the fine-grained character-word semantic space, making it difficult to distinguish the contribution of different words when a character is paired with multiple words. Specifically, as shown in Figure \ref{fig:example}, we divide the matched words into two types according to the effect of word boundaries and semantics on entity extraction: 1) \textit{Effective word}, which supports the identification of the targeted entity. 2) \textit{Disturbed word}, which is isolated from the targeted entity and conflicts with contextual understanding. Taking the truncated Chinese sequence \begin{CJK*}{UTF8}{gbsn}``北京人民大会堂\ (Beijing Great Hall of the People)'' as an example, the character ``人\ (People)'' matches to the word subset \{``北京人\ (Beijing people)'', ``人民\ (People)'', ``人民大会堂\ (Great Hall of the People)''\}. In this case, the lexicon-enhanced approaches should learn to pay more attention to effective words (``人民'' and ``人民大会堂'') and reduce the influence of disturbed words (``北京人''). Furthermore, the single-matched disturbed word injected into the corresponding characters is overly targeted, which inclines to be misidentified as an entity, such as ``公众\ (Public)''.\end{CJK*}

In this paper, we consider that it is crucial to establish dense interactions between fine-grained semantic units, which is helpful for each character to discover the latent words with high semantic correlation, and to alleviate the bias introduced by disturbed words. To this end, we propose a novel unified lattice graph fusion (ULGF) approach for Chinese NER. Inspired by the key insight from multi-modal learning~\cite{yin2020novel, liang2021graghvqa}, we attempt to regard the character sequence and lexicon as two modalities with different sources. To model dense interactions of each fine-grained semantic unit over the lattice structure, we first convert the lattice structure into a unified multi-source graph to represent the input sentence and lexicon. In the graph, each node indicates a semantic unit: latent word or character, and two types of edges are introduced to model semantic relations between fine-grained units within the same source and semantic correspondences between fine-grained units of different sources, respectively. Based on the graph, we then stack multiple graph-based multi-source fusion layers that iteratively perform semantic interactions among multi-source nodes to encode graph information. Specifically, we distinguish encoders for training characters and words, and sequentially conduct intra-source and inter-source fusions utilizing the self-attention and cross-gating mechanisms to learn multi-source node representations. Moreover, to largely eliminate the bias of disturbed words, we further introduce lexicon entity classification (LEC) as an auxiliary task.

We conduct extensive experiments on four Chinese NER benchmark datasets to evaluate the proposed approach. Experiment results show that ULGF achieves considerable improvements compared to
the state-of-the-art models.

\section{Related Work}
\paragraph{Lexicon-enhanced NER} In order to effectively avoid segmentation errors and enhance boundary representations, most recent models aimed to integrate lexicon information into character-level Chinese NER.~\cite{zhang2018chinese} first introduced the lattice LSTM to explicitly leverage the word semantic and boundary information, in which matched words are integrated into corresponding characters with the DAG structure. To overcome the disadvantages of poor parallelism and insufficient information utilization of the DAG,~\cite{gui2019cnn} proposed a CNN structure with rethinking mechanism for incorporating lexicons.~\cite{sui2019leverage} formulated the lexicons and character sequence as a graph and leveraged the graph attention network for encoding. Furthermore,~\cite{mengge2020porous} and~\cite{li2020flat} proposed Transformer-based lexical enhancement methods that can capture word information via relative position encoding.~\cite{zhao2021dynamic} designed cross-lattice and self-lattice attention modules to model the dense interactions over word-character lattice structure. Compared with previous methods, we use the graph-based and contextual interaction to model various relations between characters and words, more effectively corresponding each character to its semantic relevant word based on the context.  

\paragraph{Graph Neural Networks} Recently, graph neural networks (GNN)~\cite{gori2005new} have been proven effective in multi-modal learning by using nodes and edges to model the representation of the basis semantic units and their relations, which is helpful for feature aggregation and interaction~\cite{li2019relation}. In visual question answering,~\cite{gao2019dynamic} proposed to dynamically fuse multi-modal features with intra-modality and inter-modality information flow.~\cite{saqur2020multimodal} parsed both the images and texts into individual graphs with object entities and attributes as nodes and relations as edges, and induced a correspondence factor matrix between pairs of nodes from both modalities using message passing algorithms.~\cite{liang2021graghvqa} proposed a language-guided GNN framework to translate a question into multiple iterations of message passing among graph nodes. Inspired by the advancements in multi-modal learning, we aim to capture fine-grained semantic correlation of characters and words using a unified graph-based framework.

\begin{figure*}[t]
\begin{center}
\centerline{\includegraphics[width=0.95 \linewidth]{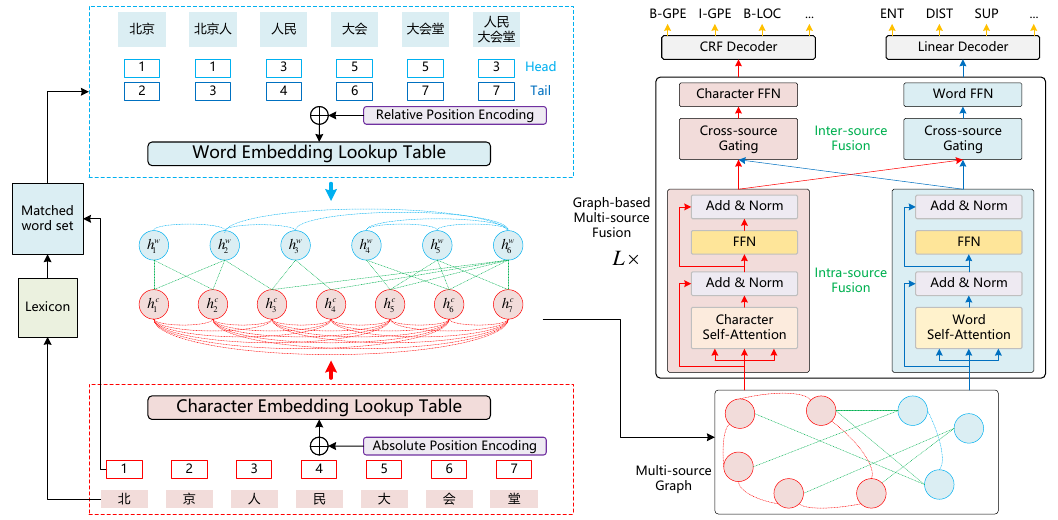}}
\vskip -0.1in
\caption{The overall architecture of ULGF.}\label{fig:3}
\end{center}
\vskip -0.4in
\end{figure*}

\section{Method}

As illustrated in Figure \ref{fig:3}, the proposed unified lattice graph fusion (ULGF) approach consists of three main components: 1) Multi-source Graph, which represents the words and character sequence in a unified space. 2) Graph-based Multi-source Fusion, which performs information interaction of different nodes through attention mechanism and adjacency relationship. 3) Training with Auxiliary Lexicon Entity Classification, which takes the attention-based contextual representation of the last hidden layer as input for multi-task learning.

\subsection{Multi-source Graph}

The graph encoder represents the input character sequence and matched word set as a unified multi-source graph. Taking the sentence in Figure \ref{fig:example} as an example, the undirected graph can be formalized as $G=(V,E)$, which is constructed as follows:

\paragraph{Node Construction}  In the node set $V$, each node represent either a character or latent word. Formally, we denote the input sentence with $n$ characters as $X_{c}=\{c_{1}, c_{2}, \ldots c_{n}\}$, where $c_{i}$ is the $i$-th character. Given a Chinese Lexicon $D$, we find out all potential words inside the sequence by matching each character with $D$. Specifically, we build a Trie based on the $D$ as~\cite{liu2021lexicon}, then match all subsequences contained in the sentence with the Trie to obtain the potential words, each of which will be bound to each character it contains. We assemble all potential words and generate a full word set $X_{w}=\{w_{1}, w_{2}, \ldots w_{m}\}$ with $m$ words for each sentence. Furthermore, We denote the word subset assigned to the $i$-th character as $ws_{i} \subseteq X_{w}$, each element of which is a word corresponding to the character.
\begin{CJK*}{UTF8}{gbsn}
For example, the word subset of the character ``人'' is $ws_{2}=$\{``北京人'', ``人民'', ``人民大会堂''\}, because these matched words can be derived from the assigned character.
\end{CJK*} 
Since the graph structure cannot directly model the order of nodes for the character sequence and word set, we inject the absolute and relative position information into the input embedding to obtain the node representations. The positional encoding with position $pos$ and dimension $k$ can be calculated as in~\citep{vaswani2017attention}:
\vskip -0.1in
\begin{equation}
\small
\renewcommand{\arraystretch}{2}
\begin{array}{c}
P_{p o s}^{(2 k)}=\sin \left(p o s / 10000^{2 k / d}\right)\\
P_{p o s}^{(2 k+1)}=\cos \left(p o s / 10000^{2 k / d}\right)\\
\end{array}
\end{equation}
For the $i$-th position in the character sequence, the character vector $h_{i}^{c} \in \mathbb{R}^{d_{c}}$ with absolute position encoding can be calculated as:
\vskip -0.1in
\begin{equation}
\small
\renewcommand{\arraystretch}{2}
\begin{array}{c}
h_{i}^{c}=\mathbf{e}^{c}\left(c_{i}\right) + P_{i}\\
\end{array}
\end{equation}
where $\mathbf{e}^{c}(\cdot)$ denotes a character embedding lookup table.

For the $j$-th word $w_j$ in the full word set $S_{w}$, We represent the positions of its head and tail characters in the character sequence as $h$ and $t$. We combine multiple distance relations and use a simple nonlinear transformation to generate a unique relative position encoding for each word. It is notable that the position of a word in $S_w$ is determined and uncorrelated with its residing word subsets. The word vector $v_{j}^{w} \in \mathbb{R}^{d_{w}}$ can be calculated as:
\vskip -0.2in
\begin{equation}
\small
\renewcommand{\arraystretch}{2}
\begin{array}{c}
R_{j}=\operatorname{ReLU}(\mathrm{W}_{r}(P_{h} \oplus P_{t} \oplus P_{t-h} \oplus P_{t+h}))\\
v_{j}^{w}=\mathbf{e}^{w}(w_{j}) + R_{j}\\
\end{array}
\end{equation}
where $\mathbf{e}^{w}(\cdot)$ denotes a pre-trained word embedding lookup table and $\mathrm{W}_{r} \in \mathbb{R}^{4d_{w} \times d_{w}}$ is a learnable matrix.

To align those two representations, we apply two multi-layer perceptrons (MLP) with Tanh activation function for the word representations to project different features onto the same space:
\begin{equation}
\small
\begin{array}{c}
h_{j}^{w}=\mathrm{W}_{2}\left(\tanh \left(\mathrm{W}_{1} v_{j}^{w}+\mathrm{b}_{1}\right)\right)+\mathrm{b}_{2}
\end{array}
\end{equation}
where $\mathrm{W}_{1}$ is a $d_{c}$-by-$d_{w}$ matrix, $\mathrm{W}_{2}$ is a $d_{c}$-by-$d_{c}$ matrix, and $\mathrm{b}_{1}, \mathrm{b}_{2}$ are scaler bias.

\paragraph{Edge Construction} To capture various semantic relations between the fine-grained multi-source units, we consider two kinds of edges in the edge set $E$: (1) Any two nodes representing characters, and any two nodes representing words assigned to the same character, are connected by an intra-source edge. (2) Each character node and all corresponding word nodes are connected by an inter-source edge. As shown in Figure \ref{fig:3}, we can observe that all characters are connected to each other, and all word nodes that are attached to the same character node are also fully-connected. Moreover, for the $i$-th character $c_{i}$ and the paired words $ws_{i}$, multiple inter-source edges are used to connect every two nodes from different sources.

\subsection{Graph-based Multi-source Fusion}
We denote the initial state of the character node sequence as $H_{c}^{(0)}=\{h_{1}^{c}, h_{2}^{c}, \ldots h_{n}^{c}\}$, and the word node set as $H_{w}^{(0)}=\{h_{1}^{w}, h_{2}^{w}, \ldots h_{m}^{w}\}$. As shown in the right part of Figure \ref{fig:3}, we stack $L$ graph-based multi-source fusion layers to encode the multi-source graph $G$. At each fusion layer, we alternately conduct intra-source and inter-source fusions to update all node states. In this way, the final node states encode both contexts within the same source and the inter-source semantic information simultaneously. Particularly, since character nodes and word nodes are two types of semantic units containing the information of different sources, we apply similar model architectures but with different parameters to model the process of state update, respectively.

Specifically, in the $l$-th fusion layer, both updates of character node states $H_{c}^{(l)}=\{h_{i}^{c}\}$ and word node states $H_{w}^{(l)}=\{h_{j}^{w}\}$ mainly involve the following steps:

\paragraph{Intra-source Fusion} We employ self-attention to generate the contextual representation of each node by collecting the information from its neighbors of the same source. We use a symmetric logical matrix to represent the connection relation of nodes. For the symmetric logical matrix $M_{c}$ to character sequence, all character nodes are full-connected, so all the elements of $M_c$ are set to 1. For the symmetric logical matrix $M_{w}$ to word set, the neighbors of each word are only those word nodes that are assigned to the common characters with it. It means that attention calculation is not carried out between non-adjacent nodes, and the corresponding elements in $M_w$ will be set to 0. We perform the multi-head self-attention by $z$ heads of attention individually and then concatenate the results. Formally, the contextual representation $T_{c}^{(l)}$ of all character nodes can be calculated as:
\vskip -0.1in
\begin{equation}
\centering
\small
\renewcommand{\arraystretch}{2}
\begin{aligned}
\operatorname{Att}_i^c=\sigma(\frac{ [H_{c}^{(l-1)}\mathrm{Q}_{i}^c] [H_{c}^{(l-1)}\mathrm{K}_{i}^c]^{T} M_{c}}{\sqrt{d_{c}}}) &[H_{c}^{(l-1)}\mathrm{V}_{i}^c]\\
T_{c}^{(l)}=[\operatorname{Att}_1^c\ldots;\operatorname{Att}_z^c]\mathrm{W}_{c}^t&
\end{aligned}
\end{equation}
where $\mathrm{Q}_{i}^c \in \mathbb{R}^{d_c \times d_z}$, $\mathrm{K}_{i}^c \in \mathbb{R}^{d_c \times d_z}$, $\mathrm{V}_{i}^c \in \mathbb{R}^{d_c \times d_z}$ denote the query, key and value parameter matrices of each head with hidden size $d_z=d_c / z$, and $\mathrm{W}_{c}^t \in \mathbb{R}^{d_c \times d_c}$ is a projection matrix for the character source. Similarly, we generate the contextual representations $T_{w}^{(l)}$ of all word nodes as:
\vskip -0.1in
\begin{equation}
\small
\begin{aligned}
\hspace{-2mm}
\operatorname{Att}_i^w=\sigma(\frac{ [H_{w}^{(l-1)}\mathrm{Q}_{i}^w] [H_{w}^{(l-1)}\mathrm{K}_{i}^w]^{T} M_{w}}{\sqrt{d_{c}}}) &[H_{w}^{(l-1)}\mathrm{V}_{i}^w]\\
T_{w}^{(l)}=[\operatorname{Att}_1^w\ldots;\operatorname{Att}_z^w]\mathrm{W}_{w}^t&
\end{aligned}
\end{equation}
where we omit the description of layer normalization and residual connection for simplicity.

\paragraph{Inter-source Fusion} Inspired by studies in multi-modal feature fusion~\cite{yin2020novel, saqur2020multimodal, yu2020improving}, we apply a cross-source gating mechanism with element-wise operation to aggregate the semantic information of the inter-source neighbours of each node.

Concretely, we generate the representation $S_{c}^{(l)}$ of a character node $c_i$ in the following way:
\begin{equation}
\centering
\small
\begin{aligned}
S_{c_{i}}^{(l)}&=T_{c_{i}}^{(l)}+\sum_{i \in \mathcal{N}(c_{i})} \alpha_{i, j} \odot T_{w_{j}}^{(l)}\\
\alpha_{i, j}=&\operatorname{sigmoid}\left(\mathrm{W}_{c 1}^{(l)} T_{c_{i}}^{(l)}+\mathrm{W}_{c 2}^{(l)} T_{w_{j}}^{(l)}\right)
\end{aligned}
\end{equation}
\vskip -0.1in
where $\mathcal{N}(c_{i})$ is the set of neighboring word nodes of $c_i$, and $\mathrm{W}_{c 1}^{(l)}$ and $\mathrm{W}_{c 2}^{(l)}$ are parameter matrices. Similarly, the representation $S_{w}^{(l)}$ of a word node $w_j$ can be calculated as follows:
\begin{equation}
\small
\begin{aligned}
S_{w_{j}}^{(l)}&=T_{w_{j}}^{(l)}+\sum_{i \in \mathcal{N}(w_{j})} \beta_{j, i} \odot T_{c_{i}}^{(l)}\\
\beta_{j, i}=&\operatorname{sigmoid}\left(\mathrm{W}_{w 1}^{(l)} T_{w_{j}}^{(l)}+\mathrm{W}_{w 2}^{(l)} T_{c_{i}}^{(l)}\right)
\end{aligned}
\end{equation}
\vskip -0.1in
where $\mathcal{N}(w_{j})$ is the set of adjacent character nodes of $w_j$, and $\mathrm{W}_{w 1}^{(l)}$ and $\mathrm{W}_{w 2}^{(l)}$ are parameter matrices.

The above fusion approach distinguishes the importance of adjacent cross-source nodes for each node by weighted aggregation, which facilitates a more precise determination of the degree of inter-source fusion according to the contextual representations of each source. Finally, we adopt position-wise feed-forward networks $\operatorname{FFN}(*)$ to generate the character nodes states $H_{c}^{(l)}$ and word node states $H_{w}^{(l)}$:
\vskip -0.1in
\begin{equation}
\small
\begin{aligned}
H_{c}^{(l)}=\operatorname{FFN}(S_{c}^{(l)})\\
H_{w}^{(l)}=\operatorname{FFN}(S_{w}^{(l)})
\end{aligned}
\end{equation}
where $S_{c}^{(l)}=\{S_{c_{i}}^{(l)}\}$ and $S_{w}^{(l)}=\{S_{w_{j}}^{(l)}\}$ denote the updated states of character and word nodes.

\subsection{Training with Auxiliary Lexicon Entity Classification}
Since character and word information has been incorporated into all node representations via multiple graph-based multi-source fusion, we take the output $H_{c}^{(L)}$ and $H_{w}^{(L)}$ of the last hidden layer for subsequent decoding.

\paragraph{NER Loss}
we feed the character representation $H_{c}^{(L)}$ into the CRF decoding layer~\cite{lafferty2001conditional} to perform sequence tagging. Let $Y^{'}$ denote the set of all arbitrary label sequences for input sentence $S_c$, the probability of the label sequence $Y$ can be calculated as:
\begin{equation}
\small
\begin{aligned}
p(Y|S_c)=\frac{\prod_{n=1}^{N}\psi_n(y_{n-1},y_n,H_{c}^{(L)})}{\sum_{y^{'}\in Y^{'}} \prod_{n=1}^{N}\psi_n(y_{n-1}^{'},y_n^{'},H_{c}^{(L)})}
\end{aligned}
\end{equation}
where $\psi_n(y_{n-1},y_n,H_{c}^{(L)})=\exp(\mathrm{W}_{crf}H_{c}^{(L)}+\mathrm{b}_{crf})$ is the potential scoring function, and $\mathrm{W}_{crf}$ and $\mathrm{b}_{crf}$ are the weight vector and bias. 

In the training phase, We optimize model parameters by minimizing the sentence-level negative log-likelihood loss as follows:
\vskip -0.1in
\begin{equation}
\small
\begin{aligned}
\mathcal{L}_{ner}=-\log p(Y|H_{c}^{(L)})
\end{aligned}
\end{equation}
\vskip -0.1in
\paragraph{LEC Loss}
Lexicon information is often integrated with positive weights, which makes disturbed words with negative influence or semantic inefficiency affect model training. To alleviate the bias, we introduce lexicon entity classification as an auxiliary training objective, to determine the common properties of matched words. As illustrated in Figure~\ref{fig:example}, according to the roles of matched words in the NER task, the LEC task automatically formulates three entity property types: \textit{Match}, \textit{Cover} and \textit{Disturb}. The word representation $H_{w}^{(L)}$ will be forwarded to a linear classification layer to predict the property type of each word. The training objective of the LEC task can be formulated as follows:
\vskip -0.2in
\begin{equation}
\small
\begin{aligned}
\mathcal{L}_{lec}=-\log softmax\left(H_{w}^{(L)} \times \mathrm{W}_{lec}+\mathrm{b}_{lec}\right)
\end{aligned}
\end{equation}
\vskip -0.1in
\paragraph{Multi-task Learning Objective}
Finally, we use a trade-off function $\lambda (t)$ to linearly combine $\mathcal{L}_{ner}$ and $\mathcal{L}_{lec}$, making the final training object as follow:
\begin{equation}
\small
\begin{aligned}
\mathcal{L}=(1-\lambda(t)) \mathcal{L}_{ner}+\lambda(t) \mathcal{L}_{lec}
\end{aligned}
\end{equation}
where $\lambda (t)$ is a two-stage function of epoch $t$: given a threshold $\tau $, when $\lambda (t)>\tau$, $\lambda (t)=\lambda_{0}\lambda_{1}^{t}$, where $\lambda_{0} \in $[$0,1$] denotes the starting value and $\lambda_{1} \in $[$0,1$] denotes the decaying value. When $\lambda (t)\leq\tau$, we will keep $\lambda (t)=\tau $.

\section{Experiments}
\subsection{Experimental Settings}
We evaluate our methods on four Chinese NER datasets, including OntoNotes~\cite{weischedel2011ontonotes}, MSRA~\cite{levow2006third}, Weibo NER~\cite{peng2015named} and Resume~\cite{zhang2018chinese}. Both OntoNotes and MSRA are in news domain, Weibo NER comes from social media data and Resume consists of resumes of senior executives. We adopt the standard Precision (P), Recall (R) and F1 score with strict match criteria as the primary metric to evaluate the models. The statistics of these datasets and implementation details are shown in Appendix. 

\begin{table*}[t]
\centering
\renewcommand{\arraystretch}{1.2}
\scalebox{0.75}{
\begin{tabular}{lccccccccccccc}
\hline
\multirow{2}{*}{\textbf{Model}}                     & \multirow{2}{*}{\textbf{Resources}}  & \multicolumn{3}{c}{\textbf{Weibo}} & \multicolumn{3}{c}{\textbf{Resume}} & \multicolumn{3}{c}{\textbf{MSRA}} & \multicolumn{3}{c}{\textbf{Ontonotes 4.0}}   \\ 
\cline{3-14}
& & \textbf{NE/P} & \textbf{NM/R} & \textbf{ F1 } & \textbf{ P } & \textbf{ R } & \textbf{ F1 } & \textbf{ P } & \textbf{ R } & \textbf{ F1 } & \textbf{ P } & \textbf{ R } & \textbf{ F1 } \\
\hline
LSTM-CRF        & - & 46.11 & 55.29 & 52.77 & 93.66 & 93.31 & 93.48 & 90.74 & 86.96 & 88.81 & 68.79 & 60.35 & 64.30 \\
Lattice-LSTM    & L & 53.04 & 62.25 & 58.79 & 94.81 & 94.11 & 94.46 & 93.57 & 92.79 & 93.18 & 76.35 & 71.56 & 73.88 \\
LR-CNN          & L & 57.14 & \textbf{66.67} & 59.92 & 95.37 & 94.84 & 95.11 & 94.50 & 92.93 & 93.71 & 76.40 & 72.60 & 74.45 \\
LGN             & L & 55.34 & 64.98 & 60.21 & 95.28 & 95.46 & 95.37 & 94.19 & 92.73 & 93.46 & 76.13 & 73.68 & 74.89 \\
FLAT            & L & -     & -     & 60.32 & -     & -     & 95.45 & -     & -     & 94.12 & -     & -     & 76.45 \\
MECT            &L+R& 61.91 & 62.51 & 63.30 & \textbf{96.40} & 95.39 & 95.89 & 94.55 & 94.09 & 94.32 & 77.57 & 76.27 & 76.92 \\
\textbf{ULGF}   & L & \textbf{64.55} & 65.29 & \textbf{65.19} & 96.37 & \textbf{96.55} & \textbf{96.46} & \textbf{95.09} & \textbf{94.82} & \textbf{94.95} & \textbf{79.02} & \textbf{79.83} & \textbf{79.42}  \\
\hline
BERT-Tagger     & - & -     & -     & 68.20 & -     & -     & 95.53 &    -    & -     & 94.95 & -     & -     & 80.14 \\
LSTM-CRF[BERT]  & - & 68.21 & 68.38 & 68.29 & 95.11 & 96.01 & 95.56 & 95.33 & 95.04 & 95.18 & 79.92 & 80.56 & 80.24 \\
FLAT[BERT]      & L & -     & -     & 68.55 & -     & -     & 95.86 &  
 -    & -     & 96.09 & -     & -     & 81.82 \\
DSCAN[BERT]     & L & -     & -     & 71.27 & -     & -     & 96.67 &    -    & -     & 96.41 & -     & -     & -     \\
MECT[BERT]      &L+R& -     & -     & 70.43 & -     & -     & 95.98 &    -    & -     & 96.24 & -     & -     & 82.57 \\
ACT-S[BERT]     & T & 72.57 & 73.95 & 73.25 & 96.30 & 96.91 & 96.60 & 96.59 & 96.89 & 96.74 & 83.98 & 83.85 & 83.91 \\
\textbf{ULGF[BERT]}      & L & \textbf{74.09} & \textbf{75.36} & \textbf{74.72} & \textbf{97.03} & \textbf{97.64} & \textbf{97.38} & \textbf{97.28} & \textbf{97.34} & \textbf{97.31} & \textbf{85.76} & \textbf{85.58} & \textbf{85.67} \\
\hline
\end{tabular}}
\vskip -0.1in

\caption{Main Results on four Chinese NER datasets with non-BERT and BERT framework. Bold marks the highest score. ‘L’ denotes the lexicon-enhanced approaches. ‘R’ denotes using Chiese radical and glyph information. ‘T’ denotes using bilingual information with translation API. The results of other models are reported in their original papers. In Weibo, NE/NM is used to evaluate non-BERT models as ~\citep{zhang2018chinese}.}
\vskip -0.2in
\label{table-experiment result-1}
\end{table*}



To verify the effectiveness of the proposed ULGF approach, we compare it to several competitive character-level baselines: LSTM-CRF~\cite{lample2016neural}, Lattice LSTM~\cite{zhang2018chinese}, LR-CNN~\cite{gui2019cnn}, LGN~\cite{gui2019lexicon}, FLAT~\cite{li2020flat} and MECT~\cite{wu2021mect}. Furthermore, to explore the effectiveness of pre-trained language models, we compare several models equipped with BERT-wmm, including common BERT-Tagger, FLAT[BERT], DSCAN~\cite{zhao2021dynamic}, MECT[BERT] and ACT-S~\cite{ning2022two}, all of which refer to replacing the original character embedding lookup table with BERT Encoder.

\subsection{Overall Performance}
Table \ref{table-experiment result-1} shows the experimental results on four Chinese NER datasets. It can be observed that the ULGF outperforms other lexicon-enhanced and radical-enhanced methods by obtaining 65.19\%, 96.46\%, 94.95\% and 79.42\% F1-value on Weibo, Resume, MSRA and OntoNotes datasets, respectively. Compared with the best results among Lattice LSTM, LR-CNN, LGN and FLAT that leverage the same character and word embeddings, ULGF gets an average F1 improvement of 3.25\%, 2.53\%, 2.34\% and 1.64\%, respectively. It indicates that by building a unified graph to capture the fine-grained semantic correlations between the character and word nodes, the ULGF can more fully exploit the semantic and boundary information of the lexicon.

Furthermore, we evaluate the models combined with BERT. We can observe that all lexicon-enhanced methods achieve higher F1-value than the baseline BERT-Tagger, which demonstrates that the word information is useful for character-level sequence tagging. Moreover, compared with the latest ACT-S model, the ULGF slightly increase by 0.57\%, 0.78\%, 1.47\% and 1.76\% on MSRA, Resume, Weibo and OntoNotes 4.0 datasets, achieving state-of-the-art performance after combining with BERT. It indicates that even though BERT model already provides rich semantic representations, ULGF consistently improves over other BERT-based models by the strong capability in exploiting lexicon information.

Furthermore, regarding the BERT-Tagging as the baseline, we can find that the model performance of ULGF[BERT] improves greatly with an average F1 of 6.02\% in the OntoNotes and Weibo datasets, while improves slightly with an average F1 of 2.11\% in the MSRA and Resume datasets. According to the data statistics in Table \ref{table-statistics}, we speculate that the reason for the significant performance improvement in these two datasets is that the entities are sparsely distributed in train set of OntoNotes, while there are fewer data and more complex label types in Weibo. In this case, the strong semantic support provided by ULGF facilitates the identification of a large number of unseen entities.

\begin{table}[t]
\centering
\small
\renewcommand{\arraystretch}{1.2}
\setlength{\tabcolsep}{3.5pt}
\begin{tabular}{p{2cm}<{\centering}p{1.1cm}<{\centering}p{1.1cm}<{\centering}p{1.1cm}<{\centering}p{1.1cm}<{\centering}}
\hline
Approaches&	Ontonotes&	MSRA&	Resume&	Weibo\\ \hline
\textbf{ULGF[BERT]}&	\textbf{85.67}&	\textbf{97.31}&	\textbf{97.38}&	\textbf{74.72}\\ \hline
WO Rel        &	84.17 &	96.53&	96.50&	73.03\\ \hline
WO Word-edge&	84.21 &	96.76&	96.79&	73.38\\ 
FC Intra-edge&	84.43 & 96.84&	96.70&	73.25\\
FC Inter-edge&	83.64 &	96.23&	96.15&	73.55\\ \hline
WO Ind     &	83.89 &	96.45&	96.22&	73.83\\\hline
\end{tabular}
\vskip -0.05in
\caption{Ablation study of our ULGF, F1 scores are evaluated on the test sets.}
\vskip -0.1in
\label{table-ablation study}
\end{table}

\begin{table}[t]
\centering
\small
\renewcommand{\arraystretch}{1.2}
\setlength{\tabcolsep}{3.5pt}
\begin{tabular}{p{2.4cm}<{\centering}p{1.4cm}<{\centering}p{1.4cm}<{\centering}p{1.4cm}<{\centering}}
\hline
Approaches&	P &	R &	F1\\ \hline
\textbf{Joint-Weight}&	\textbf{85.76}&	\textbf{85.58}&	\textbf{85.67}\\
\hline
Joint-Average&	84.27&	83.85&	84.06\\
LEC-Joint&	85.01&	84.73&	84.87\\
\hline
WO LEC&	83.89&	85.08&	84.48\\
\hline
\end{tabular}
\vskip -0.1in
\caption{Ablation study of our ULGF, F1 scores are evaluated on the test sets.}
\vskip -0.2in
\label{table-training strategy}
\end{table}

\begin{CJK*}{UTF8}{gbsn}
\begin{table*}[t]
\centering
\small
\renewcommand{\arraystretch}{1.2}
\setlength{\tabcolsep}{1.6pt}
\begin{tabular}{c|c c c c c c c c c c c c c c}
\hline
\multicolumn{15}{l}{Case 1}\\ \hline
\multirow{2}{*}{Sentence}&	\multicolumn{14}{c}{中国是联合国安理会常任理事国}\\ 
&\multicolumn{14}{c}{China is a permanent member of the UN Security Council}\\\hline
\multirow{2}{*}{Match words}&\multicolumn{14}{c}{中国，联合，联合国, 安理会, 联合国安理会, 常任，理事，常任理事国}\\ 
&\multicolumn{14}{c}{China, Unite, UN, Security Council, UN Security Council, Permanent, Council Member, Permanent Member}\\ \hline
BERT&\textcolor{blue}{B-GPE}&\textcolor{blue}{I-GPE}&O& \textcolor{red}{B-GPE}&\textcolor{red}{I-GPE}&\textcolor{red}{I-GPE}&\textcolor{red}{B-ORG}&\textcolor{red}{I-ORG}&\textcolor{red}{I-ORG}&O&O&O&O&O\\ 
FLAT[BERT]&\textcolor{blue}{B-GPE}&\textcolor{blue}{I-GPE}&O& \textcolor{blue}{B-ORG}&\textcolor{blue}{I-ORG}&\textcolor{blue}{I-ORG}&\textcolor{blue}{I-ORG}&\textcolor{blue}{I-ORG}&\textcolor{blue}{I-ORG}&\textcolor{red}{B-GPE}&\textcolor{red}{I-GPE}&\textcolor{red}{I-GPE}&\textcolor{red}{I-GPE}&\textcolor{red}{I-GPE}\\ 
ULGF[BERT]&\textcolor{blue}{B-GPE}&\textcolor{blue}{I-GPE}&O& \textcolor{blue}{B-ORG}&\textcolor{blue}{I-ORG}&\textcolor{blue}{I-ORG}&\textcolor{blue}{I-ORG}&\textcolor{blue}{I-ORG}&\textcolor{blue}{I-ORG}&O&O&O&O&O\\ \hline
\multicolumn{2}{l}{Case 2}\\ \hline
\multirow{2}{*}{Sentence}&\multicolumn{14}{c}{巴黎艺术博物馆位于红磨坊附近}\\ 
&\multicolumn{14}{c}{The Paris Art Museum is located near the Moulin Rouge}\\\hline
\multirow{2}{*}{Match words}&\multicolumn{14}{c}{巴黎, 艺术, 艺术博物馆, 博物馆, 位于, 红磨坊, 磨坊, 附近}\\ 
&\multicolumn{14}{c}{Paris, Art, Art Museum, Museum, Locate, Moulin Rouge, Mill, Nearby}\\ \hline
BERT& \textcolor{blue}{B-GPE}&\textcolor{blue}{I-GPE}&O&O&\textcolor{red}{B-ORG}&\textcolor{red}{I-ORG}&\textcolor{red}{I-ORG}&O&O& O&\textcolor{red}{B-ORG}&\textcolor{red}{I-ORG}&O&O\\ \hline
FLAT[BERT]& \textcolor{red}{B-ORG}&\textcolor{red}{I-ORG}& \textcolor{red}{I-ORG}& \textcolor{red}{I-ORG}& \textcolor{red}{I-ORG}& \textcolor{red}{I-ORG}& \textcolor{red}{I-ORG} &O& O& \textcolor{blue}{B-LOC}& \textcolor{blue}{I-LOC}& \textcolor{blue}{I-LOC}& O& O\\\hline
ULGF[BERT]&	\textcolor{blue}{B-GPE}& \textcolor{blue}{I-GPE}& \textcolor{blue}{B-ORG}& \textcolor{blue}{I-ORG}& \textcolor{blue}{I-ORG}& \textcolor{blue}{I-ORG}& \textcolor{blue}{I-ORG}& O& O& \textcolor{blue}{B-LOC}& \textcolor{blue}{I-LOC}& \textcolor{blue}{I-LOC}& O& O\\ \hline
\end{tabular}
\vskip -0.05in
\caption{The predicted results for two examples of the Weibo dataset. Contents with blue and red colors represent correct and incorrect entities, respectively.}
\vskip -0.2in
\label{table-case study}
\end{table*}
\end{CJK*}

\subsection{Ablation Study}
\paragraph{Model Architecture}
As shown in Table \ref{table-ablation study}, to investigate the importance of different factors in our ULGF, we conduct ablation experiments on all four datasets:

(1) WO Rel. Firstly, we remove the relative position encoding in the word node representation. The average F1-value of the model decreased by 1.21\%, which demonstrates the importance of relative position encoding in coupling fine-grained word and character information.

(2) Secondly, we discuss the edge construction of the unified multi-source graph, including: WO Word-edge, which removes the intra-source edges of the word nodes. FC-Intra edge and FC-Inter edge, in which any two nodes representing words are full-connected by an intra-source edge and any two nodes representing a word and a character are fully connected by an inter-source edge, respectively. The experimental results show that in all datasets, the other three edge construction strategies lead to different degrees of performance degradation, which demonstrates the superiority of our scheme. Specifically, WO Word-edge strategy lacks the intra-source interaction within matched words, while FC Intra-edge and FC Inter-edge strategies will introduce too many unrelated edges, which interfere with the semantic interaction.

(3) WO Ind. Finally, we adopt the shared parameters rather than independent encoding for the intra-source fusion module so that it leads to an obvious performance drop, which demonstrates the importance of parameter variability for encoding information about characters and words.

\begin{figure}[t]
\begin{center}
\centerline{\includegraphics[width=0.95 \linewidth]{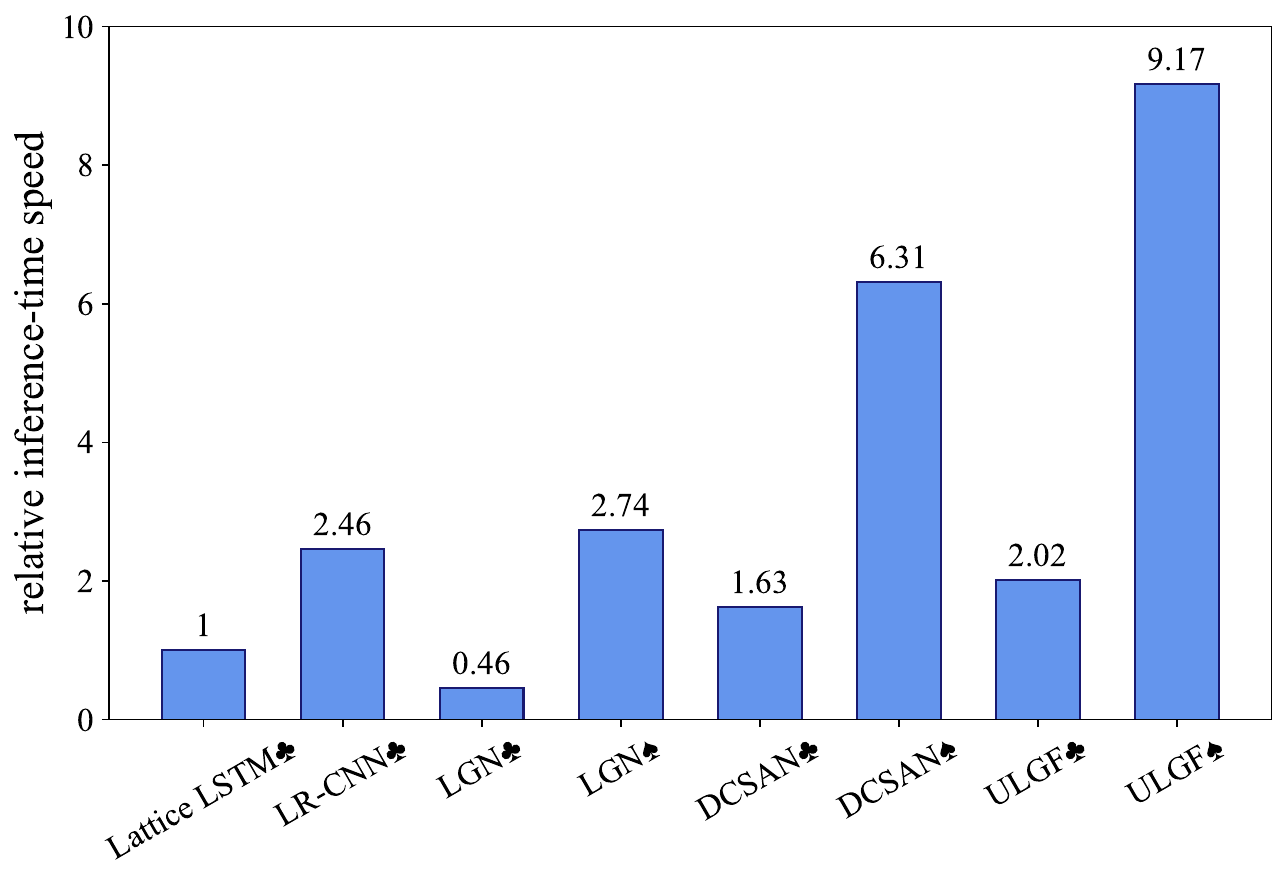}}
\vskip -0.1in
\caption{Relative inference-speed of different models compared with non-parallel Lattice LSTM$\clubsuit$, where $\clubsuit$ represents the inference speed under non-parallel conditions, and $\spadesuit$ indicates the model runs in 16 batch size parallelly.}\label{fig:efficiency}
\end{center}
\vskip -0.4in
\end{figure}

\paragraph{Training Strategy}
we investigate the influence of the multi-task learning of ULGF and use the OntoNotes dataset for illustration. We denote our training strategy as Joint-Weight and compare it with the following approaches: (1) Joint-Average, in which we directly jointly treat NER and LEC tasks with the same weight. (2) LEC-Joint, in which we freeze NER training objective to tune the LEC loss in the first five epochs, and then jointly tune both models until convergence with the same weight. (3) WO LEC, which removes LEC as an auxiliary task and fine-tunes NER loss separately.

Results are shown in Table~\ref{table-training strategy}. We can observe that the Joint-Weight outperforms the rest three strategies. It indicates that at the early stage of training, the model needs to learn the correlation and matching criteria between words and domain-specific entity types from the LEC task, and as training proceeds, we will reduce the proportion of the auxiliary task to a relatively low threshold so as not to hinder the training of the main task. Furthermore, referring to the results of the LEC-Joint strategy, we speculated the reason for the inferior performance of the Joint-Average is that the fitting of LEC task with initial parameters misleads the training process of NER task. Moreover, discarding the LEC task will lead to significant performance degradation in overall precision, while a slight increase in overall recall. The result is consistent with our hypothesis that the guidance of lexicon information drives the disturbed words to be misjudged as entities, while the LEC auxiliary task can balance the role of word and character.

\subsection{Efficiency of ULGF}
To verify the computation efficiency of our model, we conduct experiments on Ontonotes to compare the inference-speed of different lexicon-enhanced models. The result is shown in Figure \ref{fig:efficiency}. Due to the restriction of the DAG structure and the variable-sized lexical words set, Lattice LSTM and LR-CNN cannot leverage the parallel computation of GPU, while both the graph-based models LGN and ULGF can utilize batch-parallelism. However, the recurrent module of LGN degrades its inference-speed, while ULGF overcomes this limitation through the efficient parallelism of multi-head attention mechanism. Since the information encoding of both characters and words can be carried out simultaneously in the multi-source fusion module, ULGF is more efficient than the highly competitive lexicon-enhanced model DSCAN with the same attention layers. When batch size = 16, the ULGF decodes up to 3.35 times and 1.46 times faster than the LGN and DSCAN respectively, which indicates our approach benefits from batch-parallelism more significantly.


\subsection{Case Study}
To further verify the effectiveness of our ULGF in utilizing word information, we analyze the predicted results of two examples from the Weibo test set. As shown in Table \ref{table-case study}, BERT cannot directly utilize the semantic and boundary information in whole words, making it insensitive in understanding unseen entities. Therefore, \begin{CJK*}{UTF8}{gbsn}BERT wrongly identifies ``联合国\ (UN)'', 安理会\ (Security Council)'' and ``磨坊\ (Mill)'' as entities, while FLAT[BERT] and ULGF[BERT] can correct them with the help of lexicon information. Moreover, due to the flat lattice structure and relative position encoding, FLAT[BERT] fails to balance the contribution from multiple matched words. Therefore, the effect of the word ``常任理事国\ (Permanent Member)'' misleads the character sequence to be predicted as a geography entity, while the fully-connected self-attention mechanism causes disturbed information interaction between unrelated words, thus forming an error entity ``巴黎艺术博物馆\ (Paris Art Museum)''. On the contrary, the ULGF[BERT] uses the intra-source and inter-source edges to model the fine-grained character-word semantic correlation, which is beneficial for each character to better discriminate the semantically correlated words. In this way, ``巴黎\ (Paris)'' and ``艺术博物馆\ (Art Museum)'' can be effectively separated into geographical and organizational entities. Furthermore, LEC auxiliary task can explicitly restrain the effect of disturbed words, which is beneficial for fitting the role of matched words in the training process.\end{CJK*}

\section{Conclusion}
In this paper, we propose a Unified Lattice Graph Fusion approach to incorporate lexicon information in a more precise way, which converts the lattice structure into a unified graph. We treat both the words and characters as nodes and exploit various relations between multi-source semantic units. We introduce a graph-based multi-source fusion module to capture the fine-grained correlations in the word-character space and leverage the lexicon entity classification as an auxiliary task to alleviate the bias of the disturbed words. Experimental results and analysis on four benchmark NER datasets demonstrate that our model is more effective and efficient than state-of-the-art models.

\section*{Limitations}
Although our approach has made some progress in lexicon enhancement, there are still several limitations. The robust performance of our model relies on the quality of word embedding and complex self-attention and cross-gating computations. Therefore, to achieve sufficient information interaction in the fine-grained character-word space, the model training phase may cost more time (15 epochs on OntoNotes and MSRA datasets, and 40 epochs on Weibo and Resume datasets). In addition, due to the differences between the pre-trained word and BERT embeddings, we only compare with limited baselines when analyzing the performance of the ULGF. In experiments, we implement only a few comparative experiments between BERT-wwm~\cite{devlin2019bert} due to the limit of computational resources.

\section*{Ethics Statement}
Scientific work published at EMNLP 2023 must comply with the \href{https://www.aclweb.org/portal/content/acl-code-ethics}{ACL Ethics Policy}. We encourage all authors to include an explicit ethics statement on the broader impact of the work, or other ethical considerations after the conclusion but before the references. The ethics statement will not count toward the page limit (8 pages for long, 4 pages for short papers).

\section*{Acknowledgements}
This document has been adapted by Yue Zhang, Ryan Cotterell and Lea Frermann from the style files used for earlier ACL and NAACL proceedings, including those for 
ACL 2020 by Steven Bethard, Ryan Cotterell and Rui Yan,
ACL 2019 by Douwe Kiela and Ivan Vuli\'{c},
NAACL 2019 by Stephanie Lukin and Alla Roskovskaya, 
ACL 2018 by Shay Cohen, Kevin Gimpel, and Wei Lu, 
NAACL 2018 by Margaret Mitchell and Stephanie Lukin,
Bib\TeX{} suggestions for (NA)ACL 2017/2018 from Jason Eisner,
ACL 2017 by Dan Gildea and Min-Yen Kan, NAACL 2017 by Margaret Mitchell, 
ACL 2012 by Maggie Li and Michael White, 
ACL 2010 by Jing-Shin Chang and Philipp Koehn, 
ACL 2008 by Johanna D. Moore, Simone Teufel, James Allan, and Sadaoki Furui, 
ACL 2005 by Hwee Tou Ng and Kemal Oflazer, 
ACL 2002 by Eugene Charniak and Dekang Lin, 
and earlier ACL and EACL formats written by several people, including
John Chen, Henry S. Thompson and Donald Walker.
Additional elements were taken from the formatting instructions of the \emph{International Joint Conference on Artificial Intelligence} and the \emph{Conference on Computer Vision and Pattern Recognition}.

\bibliography{anthology,custom}

\begin{thebibliography}{33}
\expandafter\ifx\csname natexlab\endcsname\relax\def\natexlab#1{#1}\fi

\bibitem[{Cao et~al.(2018)Cao, Chen, Liu, Zhao, and Liu}]{cao2018adversarial}
Pengfei Cao, Yubo Chen, Kang Liu, Jun Zhao, and Shengping Liu. 2018.
\newblock Adversarial transfer learning for chinese named entity recognition
  with self-attention mechanism.
\newblock In \emph{Proceedings of the 2018 conference on empirical methods in
  natural language processing}, pages 182--192.

\bibitem[{Cui et~al.(2021)Cui, Che, Liu, Qin, and Yang}]{cui2021pre}
Yiming Cui, Wanxiang Che, Ting Liu, Bing Qin, and Ziqing Yang. 2021.
\newblock Pre-training with whole word masking for chinese bert.
\newblock \emph{IEEE/ACM Transactions on Audio, Speech, and Language
  Processing}, 29:3504--3514.

\bibitem[{Devlin et~al.(2019)Devlin, Chang, Lee, and
  Toutanova}]{devlin2019bert}
Jacob Devlin, Ming-Wei Chang, Kenton Lee, and Kristina Toutanova. 2019.
\newblock Bert: Pre-training of deep bidirectional transformers for language
  understanding.
\newblock In \emph{Proceedings of the 2019 Conference of the North American
  Chapter of the Association for Computational Linguistics: Human Language
  Technologies, Volume 1 (Long and Short Papers)}, pages 4171--4186.

\bibitem[{Gao et~al.(2019)Gao, Jiang, You, Lu, Hoi, Wang, and
  Li}]{gao2019dynamic}
Peng Gao, Zhengkai Jiang, Haoxuan You, Pan Lu, Steven~CH Hoi, Xiaogang Wang,
  and Hongsheng Li. 2019.
\newblock Dynamic fusion with intra-and inter-modality attention flow for
  visual question answering.
\newblock In \emph{Proceedings of the IEEE/CVF conference on computer vision
  and pattern recognition}, pages 6639--6648.

\bibitem[{Gori et~al.(2005)Gori, Monfardini, and Scarselli}]{gori2005new}
M~Gori, G~Monfardini, and F~Scarselli. 2005.
\newblock A new model for learning in graph domains.
\newblock In \emph{Proceedings. 2005 IEEE International Joint Conference on
  Neural Networks, 2005.}, volume~2, pages 729--734. IEEE.

\bibitem[{Gui et~al.(2019{\natexlab{a}})Gui, Ma, Zhang, Zhao, Jiang, and
  Huang}]{gui2019cnn}
Tao Gui, Ruotian Ma, Qi~Zhang, Lujun Zhao, Yu-Gang Jiang, and Xuanjing Huang.
  2019{\natexlab{a}}.
\newblock Cnn-based chinese ner with lexicon rethinking.
\newblock In \emph{ijcai}, pages 4982--4988.

\bibitem[{Gui et~al.(2019{\natexlab{b}})Gui, Zou, Zhang, Peng, Fu, Wei, and
  Huang}]{gui2019lexicon}
Tao Gui, Yicheng Zou, Qi~Zhang, Minlong Peng, Jinlan Fu, Zhongyu Wei, and
  Xuan-Jing Huang. 2019{\natexlab{b}}.
\newblock A lexicon-based graph neural network for chinese ner.
\newblock In \emph{Proceedings of the 2019 Conference on Empirical Methods in
  Natural Language Processing and the 9th International Joint Conference on
  Natural Language Processing (EMNLP-IJCNLP)}, pages 1040--1050.

\bibitem[{He and Sun(2017)}]{he2017f}
Hangfeng He and Xu~Sun. 2017.
\newblock F-score driven max margin neural network for named entity recognition
  in chinese social media.
\newblock In \emph{Proceedings of the 15th Conference of the European Chapter
  of the Association for Computational Linguistics: Volume 2, Short Papers},
  pages 713--718.

\bibitem[{Lafferty et~al.(2001)Lafferty, McCallum, and
  Pereira}]{lafferty2001conditional}
John~D Lafferty, Andrew McCallum, and Fernando~CN Pereira. 2001.
\newblock Conditional random fields: Probabilistic models for segmenting and
  labeling sequence data.
\newblock In \emph{ICML}.

\bibitem[{Lample et~al.(2016)Lample, Ballesteros, Subramanian, Kawakami, and
  Dyer}]{lample2016neural}
Guillaume Lample, Miguel Ballesteros, Sandeep Subramanian, Kazuya Kawakami, and
  Chris Dyer. 2016.
\newblock Neural architectures for named entity recognition.
\newblock In \emph{Proceedings of the 2016 Conference of the North American
  Chapter of the Association for Computational Linguistics: Human Language
  Technologies}, pages 260--270.

\bibitem[{Levow(2006)}]{levow2006third}
Gina-Anne Levow. 2006.
\newblock The third international chinese language processing bakeoff: Word
  segmentation and named entity recognition.
\newblock In \emph{Proceedings of the Fifth SIGHAN Workshop on Chinese Language
  Processing}, pages 108--117.

\bibitem[{Li et~al.(2019)Li, Gan, Cheng, and Liu}]{li2019relation}
Linjie Li, Zhe Gan, Yu~Cheng, and Jingjing Liu. 2019.
\newblock Relation-aware graph attention network for visual question answering.
\newblock In \emph{Proceedings of the IEEE/CVF international conference on
  computer vision}, pages 10313--10322.

\bibitem[{Li et~al.(2020)Li, Yan, Qiu, and Huang}]{li2020flat}
Xiaonan Li, Hang Yan, Xipeng Qiu, and Xuan-Jing Huang. 2020.
\newblock Flat: Chinese ner using flat-lattice transformer.
\newblock In \emph{Proceedings of the 58th Annual Meeting of the Association
  for Computational Linguistics}, pages 6836--6842.

\bibitem[{Liang et~al.(2021)Liang, Jiang, and Liu}]{liang2021graghvqa}
Weixin Liang, Yanhao Jiang, and Zixuan Liu. 2021.
\newblock Graghvqa: Language-guided graph neural networks for graph-based
  visual question answering.
\newblock In \emph{Proceedings of the Third Workshop on Multimodal Artificial
  Intelligence}, pages 79--86.

\bibitem[{Liu et~al.(2021)Liu, Fu, Zhang, and Xiao}]{liu2021lexicon}
Wei Liu, Xiyan Fu, Yue Zhang, and Wenming Xiao. 2021.
\newblock Lexicon enhanced chinese sequence labeling using bert adapter.
\newblock In \emph{Proceedings of the 59th Annual Meeting of the Association
  for Computational Linguistics and the 11th International Joint Conference on
  Natural Language Processing (Volume 1: Long Papers)}, pages 5847--5858.

\bibitem[{Liu et~al.(2019)Liu, Xu, Xu, Song, and Zu}]{liu2019encoding}
Wei Liu, Tongge Xu, Qinghua Xu, Jiayu Song, and Yueran Zu. 2019.
\newblock An encoding strategy based word-character lstm for chinese ner.
\newblock In \emph{Proceedings of the 2019 Conference of the North American
  Chapter of the Association for Computational Linguistics: Human Language
  Technologies, Volume 1 (Long and Short Papers)}, pages 2379--2389.

\bibitem[{Ma and Hovy(2016)}]{ma2016end}
Xuezhe Ma and Eduard Hovy. 2016.
\newblock End-to-end sequence labeling via bi-directional lstm-cnns-crf.
\newblock In \emph{Proceedings of the 54th Annual Meeting of the Association
  for Computational Linguistics (Volume 1: Long Papers)}, pages 1064--1074.

\bibitem[{Mengge et~al.(2020)Mengge, Yu, Liu, Zhang, Meng, and
  Wang}]{mengge2020porous}
Xue Mengge, Bowen Yu, Tingwen Liu, Yue Zhang, Erli Meng, and Bin Wang. 2020.
\newblock Porous lattice transformer encoder for chinese ner.
\newblock In \emph{Proceedings of the 28th International Conference on
  Computational Linguistics}, pages 3831--3841.

\bibitem[{Ning et~al.(2022)Ning, Yang, Wang, Sun, Lin, and Wang}]{ning2022two}
Jinzhong Ning, Zhihao Yang, Zhizheng Wang, Yuanyuan Sun, Hongfei Lin, and Jian
  Wang. 2022.
\newblock Two languages are better than one: Bilingual enhancement for chinese
  named entity recognition.
\newblock In \emph{Proceedings of the 29th International Conference on
  Computational Linguistics}, pages 2024--2033.

\bibitem[{Peng and Dredze(2015)}]{peng2015named}
Nanyun Peng and Mark Dredze. 2015.
\newblock Named entity recognition for chinese social media with jointly
  trained embeddings.
\newblock In \emph{Proceedings of the 2015 conference on empirical methods in
  natural language processing}, pages 548--554.

\bibitem[{Saqur and Narasimhan(2020)}]{saqur2020multimodal}
Raeid Saqur and Karthik Narasimhan. 2020.
\newblock Multimodal graph networks for compositional generalization in visual
  question answering.
\newblock \emph{Advances in Neural Information Processing Systems},
  33:3070--3081.

\bibitem[{Song et~al.(2018)Song, Shi, Li, and Zhang}]{song2018directional}
Yan Song, Shuming Shi, Jing Li, and Haisong Zhang. 2018.
\newblock Directional skip-gram: Explicitly distinguishing left and right
  context for word embeddings.
\newblock In \emph{Proceedings of the 2018 Conference of the North American
  Chapter of the Association for Computational Linguistics: Human Language
  Technologies, Volume 2 (Short Papers)}, pages 175--180.

\bibitem[{Sui et~al.(2019)Sui, Chen, Liu, Zhao, and Liu}]{sui2019leverage}
Dianbo Sui, Yubo Chen, Kang Liu, Jun Zhao, and Shengping Liu. 2019.
\newblock Leverage lexical knowledge for chinese named entity recognition via
  collaborative graph network.
\newblock In \emph{Proceedings of the 2019 Conference on Empirical Methods in
  Natural Language Processing and the 9th International Joint Conference on
  Natural Language Processing (EMNLP-IJCNLP)}, pages 3830--3840.

\bibitem[{Sun et~al.(2020)Sun, Shao, Li, Liu, Yan, Qiu, and
  Huang}]{sun2020learning}
Tianxiang Sun, Yunfan Shao, Xiaonan Li, Pengfei Liu, Hang Yan, Xipeng Qiu, and
  Xuanjing Huang. 2020.
\newblock Learning sparse sharing architectures for multiple tasks.
\newblock In \emph{Proceedings of the AAAI Conference on Artificial
  Intelligence}, volume~34, pages 8936--8943.

\bibitem[{Vaswani et~al.(2017)Vaswani, Shazeer, Parmar, Uszkoreit, Jones,
  Gomez, Kaiser, and Polosukhin}]{vaswani2017attention}
Ashish Vaswani, Noam Shazeer, Niki Parmar, Jakob Uszkoreit, Llion Jones,
  Aidan~N Gomez, {\L}ukasz Kaiser, and Illia Polosukhin. 2017.
\newblock Attention is all you need.
\newblock \emph{Advances in neural information processing systems}, 30.

\bibitem[{Weischedel et~al.(2011)Weischedel, Pradhan, Ramshaw, Palmer, Xue,
  Marcus, Taylor, Greenberg, Hovy, Belvin et~al.}]{weischedel2011ontonotes}
Ralph Weischedel, Sameer Pradhan, Lance Ramshaw, Martha Palmer, Nianwen Xue,
  Mitchell Marcus, Ann Taylor, Craig Greenberg, Eduard Hovy, Robert Belvin,
  et~al. 2011.
\newblock Ontonotes release 4.0.
\newblock \emph{LDC2011T03, Philadelphia, Penn.: Linguistic Data Consortium}.

\bibitem[{Wu et~al.(2021)Wu, Song, and Feng}]{wu2021mect}
Shuang Wu, Xiaoning Song, and Zhenhua Feng. 2021.
\newblock Mect: Multi-metadata embedding based cross-transformer for chinese
  named entity recognition.
\newblock In \emph{Proceedings of the 59th Annual Meeting of the Association
  for Computational Linguistics and the 11th International Joint Conference on
  Natural Language Processing (Volume 1: Long Papers)}, pages 1529--1539.

\bibitem[{Yang et~al.(2016)Yang, Teng, Zhang, and Zhang}]{yang2016combining}
Jie Yang, Zhiyang Teng, Meishan Zhang, and Yue Zhang. 2016.
\newblock Combining discrete and neural features for sequence labeling.
\newblock In \emph{International Conference on Intelligent Text Processing and
  Computational Linguistics}, pages 140--154. Springer.

\bibitem[{Yin et~al.(2020)Yin, Meng, Su, Zhou, Yang, Zhou, and
  Luo}]{yin2020novel}
Yongjing Yin, Fandong Meng, Jinsong Su, Chulun Zhou, Zhengyuan Yang, Jie Zhou,
  and Jiebo Luo. 2020.
\newblock A novel graph-based multi-modal fusion encoder for neural machine
  translation.
\newblock In \emph{Proceedings of the 58th Annual Meeting of the Association
  for Computational Linguistics}, pages 3025--3035.

\bibitem[{Yu et~al.(2020)Yu, Jiang, Yang, and Xia}]{yu2020improving}
Jianfei Yu, Jing Jiang, Li~Yang, and Rui Xia. 2020.
\newblock Improving multimodal named entity recognition via entity span
  detection with unified multimodal transformer.
\newblock In \emph{Proceedings of the 58th Annual Meeting of the Association
  for Computational Linguistics}, pages 3342--3352.

\bibitem[{Zhang and Yang(2018)}]{zhang2018chinese}
Yue Zhang and Jie Yang. 2018.
\newblock Chinese ner using lattice lstm.
\newblock In \emph{Proceedings of the 56th Annual Meeting of the Association
  for Computational Linguistics (Volume 1: Long Papers)}, pages 1554--1564.

\bibitem[{Zhao et~al.(2021{\natexlab{a}})Zhao, Hu, Cai, Chen, and
  Liu}]{zhao2021dynamic}
Shan Zhao, Minghao Hu, Zhiping Cai, Haiwen Chen, and Fang Liu.
  2021{\natexlab{a}}.
\newblock Dynamic modeling cross-and self-lattice attention network for chinese
  ner.
\newblock In \emph{Proceedings of the AAAI Conference on Artificial
  Intelligence}, volume~35, pages 14515--14523.

\bibitem[{Zhao et~al.(2021{\natexlab{b}})Zhao, Hu, Cai, and
  Liu}]{zhao2021modeling}
Shan Zhao, Minghao Hu, Zhiping Cai, and Fang Liu. 2021{\natexlab{b}}.
\newblock Modeling dense cross-modal interactions for joint entity-relation
  extraction.
\newblock In \emph{Proceedings of the Twenty-Ninth International Conference on
  International Joint Conferences on Artificial Intelligence}, pages
  4032--4038.

\end{thebibliography}
\bibliographystyle{acl_natbib}

\clearpage

\appendix
\section{Data Statistics}
We illustrate the train/dev/test partitions of the four Chinese NER datasets. Meanwhile, we calculate the coverage of matched words across all entities to assess the upper limits of lexicon-enhanced techniques.

\begin{table}[t]
\centering
\renewcommand{\arraystretch}{1.1}
\scalebox{0.8}{
\begin{tabular}{|l|l|l|l|l|l|}
\hline
\multicolumn{2}{|l|}{Dataset}                     & Type & Train   & Dev    & Test   \\ \hline
    \multirow{12}{*}{NER} & \multirow{3}{*}{Weibo}     & Sent & 1.4k    & 0.27k  & 0.27k  \\ \cline{3-6} 
                     &                            & Entity$_{avg}$ & 1.42  & 1.46 & 1.56 \\ \cline{3-6}
                     &                            & Rate$_{Word/Ent}$ & 81.41   & 84.31  & 82.95  \\ \cline{2-6} 
                     & \multirow{3}{*}{Ontonotes} & Sent & 15.7k   & 4.3k   & 4.3k   \\ \cline{3-6} 
                     &                            & Entity$_{avg}$ & 0.85  & 1.62 & 1.77 \\ \cline{3-6} 
                     &                            & Rate$_{Word/Ent}$ & 64.47  & 70.83 & 70.26 \\ \cline{2-6} 
                     & \multirow{3}{*}{MSRA}      & Sent & 46.4k   & -      & 4.4k   \\ \cline{3-6} 
                     &                            & Entity$_{avg}$ & 1.61  & - & 1.42 \\ \cline{3-6}
                     &                            & Rate$_{Word/Ent}$ & 69.43 & -      & 79.93 \\ \cline{2-6} 
                     & \multirow{3}{*}{Resume}    & Sent & 3.8k    & 0.46k  & 0.48k  \\ \cline{3-6} 
                     &                            & Entity$_{avg}$ & 3.52  & 3.23 & 3.42 \\ \cline{3-6} 
                     &                            & Rate$_{Word/Ent}$ & 46.94  & 60.49  & 56.81 \\ \cline{2-6} 
                     \hline
                     \end{tabular}
}
\caption{The statistics of four Chinese NER datasets. ``Sent'' is the number of sentences, ``Entity$_{avg}$'' is the average number of entities in an instance, and ``Rate$_{Word/Ent}$''(\%) is the ratio of matched words to total number of entities in a dataset.}
\vskip -0.2in
\label{table-statistics}
\end{table}

\section{Implementation Details}

We utilize the 300-dimension character embeddings as~\cite{zhang2018chinese} and word embeddings as the 200-dimension pre-trained word embedding from~\cite{song2018directional}, which is trained on free texts from news and webpages using a directional skip-gram model. For attention settings, The dimension of hidden size is 768, and the head number of multi-head self-attention in intra-source fusion are set as 8. We use the Adam optimizer with an initial learning rate of 2e-5 and the weight decay of 0.05 for parameters training. To avoid overfitting, Dropout is applied to the embeddings with a rate of 0.5 and the graph-based multi-source fusion layer with a rate of 0.3. For approaches based on BERT~\cite{devlin2019bert}, we utilize the ``BERT-wwm'' released by~\cite{cui2021pre}, and fine-tune BERT representation layer during training. We adopt standard F1 score to evaluate the models.

\end{document}